\documentclass{article}

\usepackage{microtype}
\usepackage{graphicx}
\usepackage{subcaption}
\usepackage{booktabs} 
\usepackage[table]{xcolor}
\usepackage{array}
\usepackage{enumitem}

\usepackage{hyperref}

\usepackage[accepted]{icml2026}

\usepackage{amsmath}
\usepackage{amssymb}
\usepackage{mathtools}
\usepackage{amsthm}

\usepackage[capitalize,noabbrev]{cleveref}

\usepackage{float}
\hypersetup{hidelinks}
\usepackage{xurl}
\usepackage{tikz}
\usepackage{amsfonts}
\usetikzlibrary{arrows.meta, positioning}
\usepackage[most]{tcolorbox}
\usepackage{xcolor}
\usepackage{listings}

\lstdefinestyle{promptstyle}{
  basicstyle=\ttfamily\footnotesize,
  breaklines=true,
  columns=fullflexible,
  keepspaces=true,
  showstringspaces=false,
  frame=none
}

\newtcblisting{promptblock}[2][]{
  enhanced,
  breakable,
  colback=#2!8,
  colframe=#2!65!black,
  coltitle=black,
  fonttitle=\bfseries,
  title=#1,
  listing only,
  listing style=promptstyle,
  left=1mm,
  right=1mm,
  top=1mm,
  bottom=1mm,
  boxrule=0.6pt,
  arc=2mm
}

\newcommand{\bP}{\mathbf{P}}

\newcommand{\accLow}[1]{\cellcolor{red!18}{#1}}
\newcommand{\accMidLow}[1]{\cellcolor{yellow!28}{#1}}
\newcommand{\accMid}[1]{\cellcolor{green!18}{#1}}
\newcommand{\accHigh}[1]{\cellcolor{green!38}{#1}}
\newcommand{\accTop}[1]{\cellcolor{green!62}{\textbf{#1}}}

\theoremstyle{plain}

\theoremstyle{definition}

\theoremstyle{remark}

\usepackage[disable,textsize=tiny]{todonotes}

\icmltitlerunning{PhysMent}

\begin{document}
\twocolumn[
\icmltitle{PhysMent: An Interactive Approach For LLM Reasoning In Physics Problems}

\icmlsetsymbol{equal}{$\dagger$}
\icmlsetsymbol{senior}{$\circ$}

\begin{icmlauthorlist}
\icmlauthor{Joseph Chan}{equal,inst1}
\icmlauthor{Utkarsh Jha}{equal,inst1}
\icmlauthor{Xiyin Yang}{equal,inst1}
\icmlauthor{Abhinav Jarajapu}{inst1}
\icmlauthor{Anik Sahai}{inst1}
\icmlauthor{Eddie Hu}{inst1}
\icmlauthor{Robin Jeshua Deepak}{inst1}
\icmlauthor{Stefano Saravalle}{senior,inst1}
\icmlauthor{Aditya Shah}{senior,inst1}
\end{icmlauthorlist}
\icmlaffiliation{inst1}{Algoverse AI Research}

\icmlcorrespondingauthor{Joseph Chan}{josephc3610@berkeley.edu}
\icmlcorrespondingauthor{Utkarsh Jha}{utkarsh.jha.inbox@gmail.com}
\icmlcorrespondingauthor{Xiyin Yang}{xy553@cornell.edu}

\icmlkeywords{physics reasoning, large language models, benchmarks, MuJoCo, agentic AI}

\vskip 0.3in
]



\printAffiliationsAndNotice{$^{\dagger}$Equal contribution\quad $^{\circ}$Equal senior authorship}  

\begin{abstract}
Large language models (LLMs) perform strongly on static science benchmarks, yet their ability to reason about the physical world through \emph{active experimentation} remains poorly understood. We introduce \textbf{PhysMent}, a benchmark that evaluates LLM physical reasoning via iterative, tool-mediated interaction with a MuJoCo physics simulator. Unlike static benchmarks that supply all quantities upfront, PhysMent requires models to \emph{discover} information by applying forces, querying object states, advancing time, and modifying scene geometry before answering. The benchmark comprises 105 scenes of classical mechanics, organized across four difficulty regimes (Easy/Hard $\times$ Single/Multi), three scene modalities (standard, object creation, hidden objects), and a scene-manipulation category, evaluated with a six-dimensional scoring framework. Results show that current models perform reasonably well on qualitative single-concept tasks (up to 80\% accuracy) but degrade substantially on quantitative tasks that demand precise, multi-step experimental procedures: most models fall below 30\% on the hardest single-concept category, where the bottleneck is procedural (adaptive multi-step tool use) rather than conceptual load. Across the seven models, accuracy ranges from 25\% to 67\%, with failures due to premature answer submission, inefficient exploration, and inconsistent grounding in simulator feedback rather than conceptual gaps. Our code is available \href{https://github.com/DRJCompSciWiz/PhysMent---An-Interactive-Approach-For-LLM-Reasoning-In-Physics-Problems}{\textcolor{blue}{here}}.
\end{abstract}

\section{Introduction}
\label{sec:introduction}

Large language models have reached strong performance on reading comprehension \citep{hendrycks2021measuring_mmlu}, graduate-level science \citep{rein2024gpqa}, and mathematical problem solving \citep{hendrycks2021measuring}. These capabilities are, however, fundamentally declarative: the model reads a fully specified problem and produces a text answer. Physical reasoning in the real world is different. A scientist confronting an unknown system must design experiments, apply interventions, observe outcomes, revise hypotheses, and iterate.

This distinction is what existing benchmarks miss. PIQA \citep{bisk2020piqa}, ScienceQA \citep{lu2022learn}, GPQA \citep{rein2024gpqa}, and ARC \citep{clark2018think} all test static recall of physical principles expressed in natural language. PHYRE \citep{bakhtin2019phyre} and IntPhys \citep{riochet2022intphys} involve physical environments but evaluate learned visual or RL policies, not language-mediated experimental reasoning. The closest prior work, Mind's Eye \citep{liu2022minds}, augments LLMs with a MuJoCo simulation back-end, but uses it as a \emph{one-shot oracle}: the model invokes the simulator once to retrieve a fact, rather than designing a multi-step experimental protocol. PhysMent addresses this gap directly: it requires a \emph{sustained, multi-turn experimental dialogue} where the model adaptively chooses interventions based on prior observations, contrasting fundamentally with Mind's Eye's single invocation.

We argue that a benchmark targeting this capability, using LLMs as agents for experimental physical inquiry, requires three properties: (1) \textbf{information incompleteness}: the model must choose which quantities to measure rather than reading them from the prompt; (2) \textbf{causal interventionism}: the model must apply forces, modify the scene, or advance time to observe consequent state changes; and (3) \textbf{grounded evaluation}: answers are verified by the physics engine, not a human-authored key. We stress that these are functional requirements for \emph{this} evaluation goal, not deficiencies of static benchmarks, which remain well suited to their own goal of measuring declarative physics knowledge.

We present \textbf{PhysMent}, a benchmark built on these three principles. Models are placed inside MuJoCo environments \citep{todorov2012mujoco} and given a structured API of 26 simulation tools. Each of 105 scenes presents a task, such as identifying the hollow sphere from its rolling behavior, or computing a coefficient of kinetic friction - neither of which can be solved by applying a formula alone.

\paragraph{Contributions.}
\begin{itemize}
  \item PhysMent is, to our knowledge, the first benchmark to evaluate LLMs through sustained multi-turn experimental dialogue with a 3D rigid-body simulator, in contrast to single-shot oracle invocations (Mind's Eye) or non-LLM puzzle-solving agents (PHYRE).
  \item We design 105 scenes across 13 difficulty-modality categories covering the core topics of undergraduate classical mechanics.
  \item We define a six-dimensional scoring framework that goes beyond binary accuracy to diagnose the quality of experimental strategies.
  \item We establish baseline results across frontier models and multiple prompting strategies, exposing systematic failure modes in quantitative and multi-step physical reasoning.
\end{itemize}

\section{Related Work}
\label{sec:related}

We situate PhysMent against four lines of work, each related to one facet of our contribution: physics/science reasoning benchmarks (what is measured), agentic tool use (how the model acts), LLM simulator coupling (how physics is grounded), and embodied AI (where prior interactive evaluation lives). PhysMent is distinguished by combining all four strands of assessment into a single benchmark suite. A full comparison of PhysMent to representative benchmarks can be found in Table~\ref{tab:benchmark_comparison} in Appendix~\ref{app:comparison}.

\subsection{Physics and Science Reasoning Benchmarks}

\paragraph{Static text-based benchmarks.}
PIQA \citep{bisk2020piqa} tests physical commonsense with $\sim$20k binary-choice questions; every problem is self-contained and requires no interaction. ScienceQA \citep{lu2022learn} and ARC \citep{clark2018think} cover elementary-to-high-school science in multiple-choice format. GPQA \citep{rein2024gpqa} extends the difficulty to expert-written, graduate-level questions in physics, biology, and chemistry that are resistant to lookup-based cheating, yet still present fully-specified problems requiring no experimental intervention. MATH \citep{hendrycks2021measuring} and MMLU \citep{hendrycks2021measuring_mmlu} evaluate competition mathematics and broad multitask science knowledge respectively. A recent wave of physics-specific benchmarks sharpens this paradigm. PhysReason \citep{zhang2025physreason} introduces step-level evaluation across physics theorem application, process understanding, calculation, and condition analysis; PHYBench \citep{qiu2025phybench} curates 500 problems and proposes a graded Expression Edit Distance score in place of binary correctness; and UGPhysics \citep{xu2025ugphysics} and ABench-Physics \citep{zhang2025abench} target contamination-resistant undergraduate physics. All of these remain static: a fully specified problem is presented and the model must derive the answer. None of these benchmarks tests the ability to \emph{design} measurements or reason from \emph{observed} experimental outcomes.

\paragraph{Vision-based physical understanding.}
PhysBench \citep{chow2025physbench} evaluates vision-language models on physical world understanding using over 10,000 video- and image-based entries, finding significant gaps across 75 VLMs. It evaluates \emph{perception} of pre-recorded events, not active interaction with a live simulator. IntPhys \citep{riochet2022intphys} uses a violation-of-expectation paradigm from developmental psychology to test intuitive physics from rendered video; CoPhy \citep{baradel2020cophy} targets counterfactual prediction of physical dynamics from visual input. Both evaluate passive prediction of pre-recorded physics — they do not permit agents to intervene in, modify, or query a live simulation.

\paragraph{Interactive physical environments.}
PHYRE \citep{bakhtin2019phyre} is a 2D classical mechanics puzzle benchmark where a learned agent places objects to satisfy a goal. While interactive, it employs simplified 2D physics, does not evaluate LLMs, and does not require language-mediated reasoning about physical quantities. ScienceWorld \citep{wang2022scienceworld} tests agents on elementary science tasks in a text-based interactive environment but uses a rule-based text simulator rather than a physics engine.

\paragraph{Mind's Eye and UTOPIA.}
Mind's Eye \citep{liu2022minds} is the most closely related prior work. It grounds LLM reasoning in physics by generating simulation code from text, executing it in MuJoCo, and feeding the output back as additional context. It introduces UTOPIA, a 39-subtask benchmark across 6 physical scenes (motion, friction, free fall, horizontal projectile, elastic collision, and inclined plane), where each subtask is a fixed observed/queried-concept pair. As established in Section~\ref{sec:introduction}, PhysMent differs from Mind's Eye in this fundamental respect: Mind's Eye invokes the simulator \emph{once} as an oracle to retrieve a fact, whereas PhysMent requires a sustained multi-turn protocol in which interventions are chosen adaptively. A single-invocation oracle has no notion of iteration budget, exploration efficiency, or grounding across turns. In PhysMent, we subsume UTOPIA's physical scope while adding $2.7\times$ more scenes, object-creation and hidden-object modalities absent in UTOPIA, and a broader 26-tool interactive API (vs.\ code generation).

\subsection{Agentic LLMs and Tool-Augmented Reasoning}

PhysMent places tool-augmented reasoning inside a physics simulation loop
where tools have physically grounded semantics and answers are verified
against the MuJoCo simulator's internal state. Prior work on tool-augmented language
models has shown that interleaving reasoning, planning, and external tool
use can substantially improve interactive problem solving.
ReAct \citep{yao2023react} interleaves verbal reasoning traces with
concrete tool actions, outperforming chain-of-thought and imitation
baselines by wide margins on interactive benchmarks such as ALFWorld
\citep{shridhar2021alfworld}. Toolformer \citep{schick2024toolformer}
teaches LMs to self-supervise API call generation, enabling zero-shot
use of calculators, search engines, and calendars. HuggingGPT
\citep{shen2023hugginggpt} demonstrates that a large LLM can
orchestrate specialist models through a planning-selection-execution
pipeline. DEPS \citep{wang2023deps} uses structured description and
self-explanation for zero-shot multi-task planning in Minecraft;
Voyager \citep{wang2023voyager} extends this to lifelong skill
acquisition.

PAL \citep{gao2023pal} offloads computation to a Python interpreter
via program-aided reasoning chains. Chameleon \citep{lu2023chameleon}
composes heterogeneous tools under LLM orchestration.
Chain-of-thought prompting \citep{wei2022chain} provides the
foundational technique for eliciting multi-step reasoning via
exemplar demonstrations. PhysMent inherits the interleaved reason-act loop of these methods but constrains the tool space to physically grounded operations whose outputs are dictated by a simulator rather than retrieved text. As such, tool feedback in PhysMent carries verifiable physical semantics.

\subsection{LLMs Coupled with Physics Simulators}

Code as Policies \citep{liang2023code} and ProgPrompt
\citep{singh2023progprompt} show that LLMs can generate executable
robot task plans interfacing with physical APIs. Inner Monologue
\citep{huang2023inner} improves long-horizon robot task completion by
closing the reasoning loop with natural language environment feedback.
Language to Rewards \citep{yu2023language} couples GPT-4 with MuJoCo
MPC to synthesize robotic skills; Eureka \citep{ma2024eureka} uses
evolutionary reward-code optimization with Isaac Gym, surpassing
expert-designed rewards on 83\% of tasks. RT-2 \citep{zitkovich2023rt2}
co-fine-tunes a vision-language model on robot trajectory data for
emergent physical control. SayCan \citep{ahn2022saycan} grounds LLM
outputs in robotic affordance functions for long-horizon task execution.

These systems integrate LLMs with physics engines for \emph{robotic
control}. PhysMent repurposes the LLM by treating the simulator as a measurement tool, using it as a reliable source of ground truth. The focus isn’t on controlling anything, but on observing how the LLM behaves over multiple steps as it interacts with a 26-tool API.

\subsection{Embodied AI Benchmarks}

EmbodiedQA \citep{das2018embodied} requires navigation and question
answering in 3D environments. ALFRED \citep{shridhar2020alfred}
evaluates language grounding in household task execution. ALFWorld
\citep{shridhar2021alfworld} aligns text-based and visually embodied
environments to study policy transfer. Habitat \citep{savva2019habitat}
and AI2-THOR \citep{kolve2017ai2thor} provide high-fidelity 3D
platforms for visual navigation and manipulation. BEHAVIOR
\citep{srivastava2021behaviorbenchmarkeverydayhousehold} covers 100 household activities with
rich physical state changes; MineDojo \citep{fan2022minedojo} builds
an open-ended agent framework on internet-scale Minecraft knowledge.
These benchmarks target general embodied acting and do not provide
access to the underlying physical quantities such as forces, momenta, torques,
energies that are central to PhysMent.

\section{Method}
\label{sec:method}

\subsection{Motivation}
Existing benchmarks for evaluating physical reasoning in LLMs rely predominantly on answering questions based on text, and all relevant information is provided in the prompt. While such evaluations measure a model's ability to recall and apply physics formulas, they fail to capture a fundamental aspect of scientific reasoning: the capacity to design experiments, gather evidence through interaction, and iteratively refine hypotheses based on observed outcomes. In physics problem solving, the investigator often lacks immediate access to all necessary quantities and must instead probe the system to extract them. PhysMent evaluates LLMs under these conditions, where the model must act as an agent within a physical environment, deciding what to measure, when to intervene, and how to synthesize observations into a final answer.

\subsection{Pipeline Formulation}

We define the PhysMent evaluation pipeline as follows. Let $\mathcal{S} = \{s_1, s_2, \dots, s_N\}$ denote a set of $N$ physics scenes, where each scene $s_i$ is specified by an XML configuration file $\mathcal{X}_i \in \mathcal{L}_{\text{XML}}$ that defines the physical world, and a metadata file $\mathcal{M}_i$ that encodes the task description $\tau_i$, the set of observable object identifiers $\mathcal{O}_i = \{o_1, o_2, \dots, o_{K_i}\}$, a permission matrix $\mathbf{P}_i \in \{0,1\}^{K_i \times D}$ governing which of $D$ physical attributes (e.g., mass, position, geometry type) are accessible per object, and the ground-truth answer $a_i^*$.

The simulator $\Phi$ maintains a physical state $\mathbf{x}_t \in \mathbb{R}^m$ at each timestep $t$, where $m$ represents the positions, velocities, and orientations of all bodies. The simulator changes according to:
\begin{equation}
    \mathbf{x}_{t+1} = \Phi(\mathbf{x}_t, \mathbf{u}_t)
\end{equation}
where $\mathbf{u}_t \in \mathcal{U}$ is an action drawn from the action space $\mathcal{U}$. The action space comprises a discrete set of tool calls $\mathcal{T} = \{f_1, f_2, \dots, f_T\}$, where each tool $f_j$ maps from a parameter space to either a state modification or an observation:
\begin{equation}
    f_j : \Theta_j \rightarrow \mathcal{A} \cup \mathcal{R}
\end{equation}
Here $\Theta_j$ denotes the parameter space of tool $f_j$, $\mathcal{A}$ denotes the set of state-modifying actions, and $\mathcal{R}$ denotes the space of returned observations. Available tools include state queries (e.g., \texttt{get\_position}, \texttt{get\_velocity}), state modifications (e.g., \texttt{apply\_force}, \texttt{move\_object}, \texttt{step}), and a terminal action \texttt{answer} that submits the model's response.

An LLM agent $\pi$ interacts with the simulator over a bounded number of iterations $I_{\max}$. At each iteration $k \in \{1, \dots, I_{\max}\}$, the agent receives the cumulative context $\mathbf{c}_k$ consisting of the initial prompt (derived from $\tau_i$, $\mathcal{O}_i$, and $\mathbf{P}_i$) and all prior tool calls and their returned results. The agent then produces a response $r_k$ containing reasoning in natural language and a sequence of tool calls:
\begin{equation}
    r_k = \pi(\mathbf{c}_k), \quad r_k = (\text{reasoning}_k, \, [f_{j_1}(\theta_1), \, f_{j_2}(\theta_2), \, \dots])
\end{equation}
The tool calls are executed sequentially by the simulator, and the returned observations are appended to form $\mathbf{c}_{k+1}$. The interaction terminates when either the agent calls the \texttt{answer} tool with its prediction $\hat{a}_i$, or $k = I_{\max}$ is reached. In the latter case, the episode is scored as incorrect regardless of any partial reasoning produced.

\subsection{Scene Taxonomy \& Design}

Each scene $s_i$ is assigned a problem type $p_i$ drawn from the set \{\texttt{comparison}, \texttt{calculation}, \texttt{computation}, \texttt{observation}\} that determines both the expected answer format and the evaluation criteria. Comparison problems require the agent to identify which object satisfies a given physical property (e.g., which sphere is hollow). Calculation and computation problems require the agent to derive a numerical quantity from simulator interaction (e.g., coefficient of friction, angular velocity). Observation problems require the agent to describe a qualitative physical phenomenon.

The scenes cover core topics in classical mechanics including kinematics, Newton's laws, rotational dynamics, energy conservation, momentum conservation, oscillatory motion, and friction. Objects in each scene are parameterized by physical attributes such as mass $m \in \mathbb{R}^+$, position $\mathbf{p} \in \mathbb{R}^3$, orientation $\mathbf{q} \in \mathbb{H}$ (unit quaternions), geometry type, density $\rho \in \mathbb{R}^+$, and friction coefficients $\boldsymbol{\mu} \in \mathbb{R}^3$. Table~\ref{tab:taxonomy} summarizes the full scene taxonomy across 105 scenes.

To make the taxonomy reproducible, we define each axis by an explicit construction rule applied at design time. \textbf{Single vs.\ Multi} is the number of distinct physical concepts whose correct application is required to reach the answer: \emph{Single} scenes require exactly one concept (e.g., projectile kinematics), while \emph{Multi} scenes require composing $\geq 2$ concepts (e.g., friction followed by energy conservation). \textbf{Easy vs.\ Hard} is determined by the minimum number of distinct tool-call \emph{types} and simulator interventions a reference solution requires: \emph{Easy} scenes are solvable with a short fixed measurement sequence ($\leq 3$ intervention steps, e.g., displace--step--read), whereas \emph{Hard} scenes require an experimental loop in which a later action depends on an earlier observation ($> 3$ steps, e.g., a binary search over incline angle to find the sliding threshold). \textbf{Object Creation} scenes require the agent to instantiate at least one new body via \texttt{create\_objects} to construct a controlled comparison (e.g., adding a reference mass), so the answer is unreachable from the initial scene alone. \textbf{Hidden Objects} scenes contain at least one task-relevant body that is not exposed in the initial object list or is visually/parametrically masked, and must be revealed through \texttt{find\_objects} before it can be measured. Per-axis definitions, a worked example for every cell, and the reference solution length used to assign Easy/Hard are provided in Appendix~\ref{app:scenes}, Table~\ref{tab:category_defs}.

We note that \emph{Hard Multi} scenes are not uniformly harder in measured accuracy than \emph{Easy Multi} (Section~\ref{sec:results}). This is by design and reflects the two orthogonal axes above: \emph{Easy/Hard} grades the length and adaptivity of the required \emph{experimental procedure}, while \emph{Single/Multi} grades \emph{conceptual} load. We make this decoupling explicit because it is itself a finding: procedural difficulty and conceptual difficulty are distinct failure axes for current models (Section~\ref{sec:discussion}).

All scenes run under MuJoCo 3.3.3 \citep{todorov2012mujoco} with a
fixed timestep of $0.005$\,s. We adopt MuJoCo for four reasons that bear directly on benchmark validity: deterministic rigid-body dynamics (reproducible ground truth), programmatic access to internal state (engine-verifiable answers), fixed-timestep control (well-defined \texttt{step} tool semantics), and direct comparability with Mind's Eye \citep{liu2022minds}. Detailed information regarding the 26 tools in PhysMent can be found in Table~\ref{tab:tools} in Appendix~\ref{app:tools}.






\begin{table*}[t]
\centering
\caption{PhysMent scene taxonomy (105 scenes). Scene IDs and folder names are
         \textbf{consecutive} $7$--$111$ (\texttt{Scenes/Scene\{N\}/scene\{N\}.json}).}
\label{tab:taxonomy}
\footnotesize
\begin{tabular}{llcc}
\toprule
\textbf{Scene ID} & \textbf{Category} & \textbf{N} & \textbf{Primary Topics} \\
\midrule
7--15   & Easy Single Physics           &  9 & Kinematics, free-fall, inclined planes \\
16--20  & Easy Single + Object Creation &  5 & Same + object manipulation \\
21--25  & Easy Single + Hidden Objects  &  5 & Same + object discovery \\
26--40  & Hard Single Physics           & 15 & Ladders, equilibrium, torque, angular momentum \\
41--45  & Hard Single + Object Creation &  5 & Hard + object manipulation \\
46--50  & Hard Single + Hidden Objects  &  5 & Hard + object discovery \\
51--65  & Easy Multi-Physics            & 15 & Combined concept problems \\
66--70  & Easy Multi + Object Creation  &  5 & Multi-concept + object creation \\
71--75  & Easy Multi + Hidden Objects   &  5 & Multi-concept + discovery \\
76--90  & Hard Multi-Physics            & 15 & Complex multi-concept \\
91--95  & Hard Multi + Object Creation  &  5 & Complex + object creation \\
96--100 & Hard Multi + Hidden Objects   &  5 & Complex + discovery \\
101--111& Create / Delete / Attach      & 11 & Dynamic scene manipulation \\
\bottomrule
\end{tabular}
\end{table*}
\subsection{Evaluation}

Given a predicted answer $\hat{a}_i$ and the ground-truth $a_i^*$, a response is marked correct if one of the following holds: (1) both answers are numerical and lie within a mixed absolute--relative tolerance, or (2) the predicted answer is contained in the ground-truth. Formally:
\begin{equation}
    C(s_i) = 
    \begin{cases}
        1 & \text{\small if } |\hat{a}_i - a_i^*| \leq \epsilon_{\text{abs}} + \epsilon_{\text{rel}}\,|a_i^*| \\
        1 & \text{\small if } \hat{a}_i \subseteq a_i^* \\
        0 & \text{\small otherwise}
    \end{cases}
    \label{eq:correctness}
\end{equation}
We set $\epsilon_{\text{rel}} = 0.05$ (5\% relative tolerance) to accommodate floating-point accumulation across multi-step tool calls, together with a small absolute floor $\epsilon_{\text{abs}}$ (in the units of the queried quantity) so that the criterion remains well defined when $a_i^* = 0$; pure relative tolerance would otherwise divide by zero for ground-truth-zero answers (e.g., a net force of $0$\,N at equilibrium). This mixed form follows the standard \texttt{isclose} convention \citep{pep485}. For categorical and comparison answers, exact string matching applies after normalization to lowercase (see Section~\ref{sec:normalization} for the full answer-normalization protocol).
The overall accuracy for an agent $\pi$ across all $N$ scenes is:
\begin{equation}
    \text{Acc}(\pi) = \frac{1}{N} \sum_{i=1}^{N} C(s_i)
\end{equation}

\subsection{Scoring Framework}

Beyond binary correctness, each scene yields a six-dimensional score
normalized to $[0,1]$ and combined into a weighted Final Score:
Let $\mathbf{w} = (w_C, w_E, w_G, w_V, w_R, w_{G'})$ with $\sum_j w_j = 1$. Then:
\begin{equation}
  \mathrm{FS}_i = 100\,\mathbf{w}^\top (C, E, G, V, R, G')^\top
\end{equation}
The dimensions are defined as follows. \textbf{Correctness} ($C$) is the per-scene correctness from Eq.~\eqref{eq:correctness} (binary at the strict threshold; see Section~\ref{sec:normalization} for the graded variant). \textbf{Efficiency} ($E$) is the budget utilization $1 - k/I_{\max}$, where $k$ is the iteration on which \texttt{answer} is called. We report $E$ as a descriptive utilization measure rather than a quality signal: because it rewards early submission, it can in principle penalize an agent that spends additional iterations verifying a correct answer, a tension we discuss in Section~\ref{sec:discussion} and which motivates reporting $E$ separately rather than folding it silently into a single score. \textbf{Groundedness} ($G$) is the fraction of quantitative assertions in the model's reasoning that are traceable to a value returned by a prior tool call, following the faithfulness/attribution tradition in retrieval-augmented and agentic evaluation \citep{es2024ragas}. \textbf{Action Validity} ($V$) is the fraction of tool calls returning valid (non-error) results. \textbf{Reasoning Diversity} ($R$) is $1 -$ maximum bigram repetition across the reasoning text; this is a deliberately lightweight \emph{lexical} proxy in the spirit of distinct-$n$ \citep{li2016diversity}, intended to flag agents that loop on identical tool-call phrasings. \textbf{Generalization} ($G'$) is held-out category accuracy.
We set $\mathbf{w} = (0.4, 0.15, 0.15, 0.10, 0.10, 0.10)$ for $(w_C, w_E, w_G, w_V, w_R, w_{G'})$ respectively, weighting correctness the heaviest. Full per-dimension results are reported in Appendix~\ref{app:dimensions}.

\subsection{Design Hypotheses and Experimental Conditions}

We believe that model performance will strongly correlate with the problem complexity. Specifically, we expect models to perform well on comparison and observation tasks that primarily require qualitative reasoning, such as identifying which object is heavier or which sphere is hollow based on rolling behavior. On the other hand, we expect significantly lower performance on calculation and computation tasks that require calculations, multi-step tool usage, and the correct application of physical formulas. Furthermore, we predict that the bounded iteration count $I_{\max}$ will be a constraint, as models that fail to plan their tool usage efficiently are likely to use up their iterations before submitting an answer, resulting in a failure mode that is different from incorrect reasoning. We also want to see the effect of iteration budget, so we will evaluate each agent across a range of iteration counts $I_{\max} \in \{5, 10, 15, 20\}$ and report accuracy as a function of $I_{\max}$, identifying the point at which additional iterations yield diminishing returns. The agent is explicitly told its $I_{\max}$ in the prompt, so planning is part of the task rather than a hidden constraint. The four budgets were chosen to bracket the regime of interest: $I_{\max}=5$ is near the minimum that allows a measure--intervene--remeasure loop for our Easy scenes, $I_{\max}=20$ is where added iterations stop changing accuracy for every model we tested (Section~\ref{sec:results}), and the intermediate values resolve the shape of the curve in between. The design of PhysMent is that it is a diagnostic tool that can help identify different failure reasons in physical reasoning.

\section{Results}
\label{sec:results}

\begin{figure*}[t]
\centering
\includegraphics[width=\textwidth]{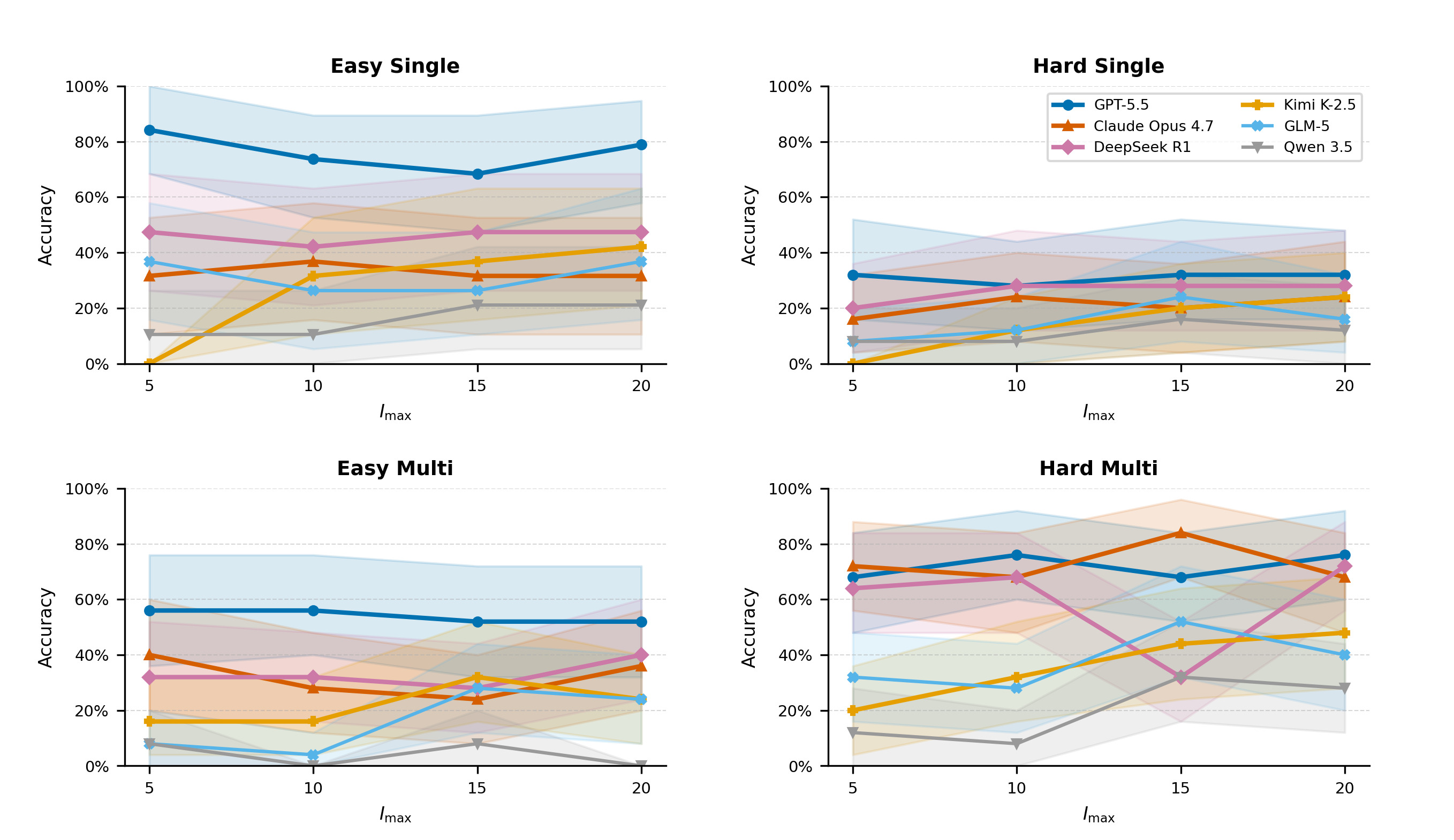}
\caption{Accuracy vs.\ iteration budget $I_{\max}$ across the four major scene categories. Models show three distinct patterns: steady improvement (Gemini 3.1 Pro, Kimi K-2.5), marginal decline with more iterations (GPT-5.5, Claude Opus 4.7), and plateauing (GLM-5, DeepSeek R1, Qwen 3.5).}
\label{fig:accuracy_vs_imax}
\end{figure*}

\begin{figure*}[t]
\centering
\scriptsize
\setlength{\tabcolsep}{3.5pt}
\renewcommand{\arraystretch}{1.18}
\begin{tabular}{p{2.7cm}ccccccc}
\toprule
\textbf{Category} & \textbf{GPT-5.5} & \textbf{Gemini} & \textbf{Opus} & \textbf{DSR1} & \textbf{Qwen} & \textbf{Kimi} & \textbf{GLM} \\
\midrule
Easy Single Phys.     & \accTop{80.6}    & \accHigh{72.2}   & \accMidLow{25.0} & \accHigh{66.7}   & \accMidLow{33.3} & \accMid{50.0}    & \accHigh{61.1} \\
Easy Single + Obj.C.  & \accMid{50.0}    & \accMid{55.0}    & \accMid{45.0}    & \accMid{40.0}    & \accLow{5.0}     & \accMidLow{20.0} & \accMidLow{25.0} \\
Easy Single + Hid.    & \accHigh{60.0}   & \accMid{50.0}    & \accMidLow{25.0} & \accMid{45.0}    & \accMidLow{20.0} & \accMidLow{30.0} & \accMid{45.0} \\
Hard Single Phys.     & \accMidLow{36.7} & \accLow{15.0}    & \accMidLow{25.0} & \accLow{6.7}     & \accLow{1.7}     & \accLow{5.0}     & \accLow{1.7} \\
Hard Single + Obj.C.  & \accLow{0.0}     & \accMid{50.0}    & \accLow{0.0}     & \accMidLow{25.0} & \accMidLow{30.0} & \accMidLow{30.0} & \accMidLow{20.0} \\
Hard Single + Hid.    & \accMid{40.0}    & \accMid{40.0}    & \accMidLow{30.0} & \accMid{40.0}    & \accLow{0.0}     & \accMidLow{20.0} & \accMidLow{35.0} \\
Easy Multi            & \accMid{41.7}    & \accMid{45.0}    & \accMidLow{21.7} & \accMidLow{31.7} & \accLow{5.0}     & \accMidLow{30.0} & \accMidLow{25.0} \\
Easy Multi + Obj.C.   & \accMid{40.0}    & \accMid{50.0}    & \accHigh{65.0}   & \accMidLow{25.0} & \accLow{10.0}    & \accMidLow{35.0} & \accMidLow{35.0} \\
Easy Multi + Hid.     & \accTop{85.0}    & \accTop{80.0}    & \accMidLow{30.0} & \accHigh{60.0}   & \accMidLow{25.0} & \accMid{55.0}    & \accMid{50.0} \\
Hard Multi            & \accHigh{66.7}   & \accMid{58.3}    & \accHigh{71.7}   & \accMidLow{31.7} & \accMidLow{23.3} & \accMid{40.0}    & \accMid{46.7} \\
Hard Multi + Obj.C.   & \accTop{80.0}    & \accHigh{70.0}   & \accMid{50.0}    & \accMidLow{25.0} & \accLow{10.0}    & \accMidLow{25.0} & \accMidLow{30.0} \\
Hard Multi + Hid.     & \accHigh{60.0}   & \accMid{50.0}    & \accTop{100.0}   & \accMidLow{25.0} & \accLow{5.0}     & \accLow{15.0}    & \accMidLow{25.0} \\
Create / Del. / Att.  & \accLow{18.2}    & \accMid{56.8}    & \accMidLow{38.6} & \accMidLow{27.3} & \accLow{13.6}    & \accMidLow{25.0} & \accLow{18.2} \\
\bottomrule
\end{tabular}

\vspace{0.35em}
\begin{tabular}{ccccc}
\accLow{$<20$} & \accMidLow{$20$--$39$} & \accMid{$40$--$59$} & \accHigh{$60$--$79$} & \accTop{$\geq80$}
\end{tabular}

\caption{Per-category accuracy heatmap across models, averaged over iteration budgets. Cell colors indicate accuracy buckets: red for $<20\%$, yellow for $20$--$39\%$, light green for $40$--$59\%$, mid green for $60$--$79\%$, and dark green for $\geq80\%$. Several modality cells contain only 5 scenes (Table~\ref{tab:taxonomy}), so a single scene shifts that cell's accuracy by roughly 20 percentage points; per-cell values should be read as coarse indicators rather than precise estimates.}
\label{fig:category_heatmap}
\end{figure*}

\subsection{Model Performance}

\begin{table*}[!t]
\centering
\caption{Overall accuracy (\%) of each model on PhysMent across iteration budgets.}
\label{tab:results_overall}
\renewcommand{\arraystretch}{0.90}
\begin{tabular}{l c c c c c c c}
\toprule
$I_{\max}$ & Gemini 3.1 Pro & GPT-5.5 & Claude Opus 4.7 & Kimi K-2.5 & GLM-5 & DeepSeek R1 & Qwen 3.5 \\
\midrule
5  & 29.5\% & \textbf{51.4\%} & 40.5\% & 27.4\% & 46.7\% & 42.3\% & 26.7\% \\
10 & 50.4\% & 49.5\% & 39.6\% & 37.1\% & 37.7\% & \textbf{51.4\%} & 24.8\% \\
15 & \textbf{51.4\%} & 48.6\% & 42.3\% & 46.7\% & 48.6\% & 48.6\% & 35.2\% \\
20 & \textbf{66.7\%} & 50.5\% & 40.5\% & 56.2\% & 48.6\% & 53.3\% & 27.6\% \\
\bottomrule
\end{tabular}
\end{table*}

Table~\ref{tab:results_overall} shows the overall accuracy across all seven models and four iteration budgets. GPT-5.5 achieves its highest accuracy at 51.4\% under $I_{\max}=5$, while Gemini 3.1 Pro reaches its peak of 66.7\% at $I_{\max}=20$. DeepSeek R1 and GLM-5 fluctuate within the roughly 38--53\% range across budgets. Gemini 3.1 Pro shows the largest improvement with iterations, rising from 29.5\% to 66.7\%, with Kimi K-2.5 close behind (27.4\% to 56.2\%). Qwen 3.5 consistently scores lowest, remaining between 24.8\% and 35.2\% across all budgets.

Figure~\ref{fig:accuracy_vs_imax} illustrates these different trajectories across the four major category groups, showing that the marginal decline pattern for GPT-5.5 and Claude Opus 4.7 is most noticeable in Easy Single and Easy Multi scenes.

Table~\ref{tab:main_results} shows per-model metrics averaged across iteration budgets. GPT-5.5 has the highest accuracy (50.0\%), while GLM-5 achieves the highest groundedness score (0.744) despite ranking fourth in accuracy. This shows that GLM-5 grounds its reasoning in simulator observations more consistently than it converts that grounding into correct answers. DeepSeek R1 shows the opposite pattern: high accuracy (48.5\%) with the lowest groundedness (0.514), indicating heavier reliance on prior knowledge rather than obtaining results from the simulator.

Figure~\ref{fig:category_heatmap} breaks down accuracy by scene category and presents a heatmap. Easy Single Physics scenes are the strongest category overall, with GPT-5.5 reaching 80.6\% and Gemini 3.1 Pro reaching 72.2\%. Hard Single Physics is the hardest category despite involving only one physical concept, with most models scoring below 40\%, which shows that precision in multi-step tool usage is the bottleneck rather than conceptual complexity. Easy Multi + Hidden scenes surprisingly outperform their non-hidden counterparts, with Claude Opus 4.7 reaching 30.0\% and Gemini 80.0\%, indicating that the \texttt{find\_objects} tool is reliably invoked once models recognize it is needed.

\section{Discussion}
\label{sec:discussion}
\subsection{The Effect of Iteration Budget}
Contrary to the expectation that more iterations always help, our results show three distinct patterns across models. First, some models improve steadily with more iterations: Gemini 3.1 Pro more than doubled its accuracy (29.5\% $\to$ 66.7\%), and Kimi K-2.5 similarly doubled (27.4\% $\to$ 56.2\%). Second, some models showed a gradual decrease in accuracy: GPT-5.5 dropped slightly from 51.4\% at $I_{\max}=5$ to 48.6\% at $I_{\max}=15$, and Claude Opus 4.7 similarly remained flat with a slight downward trend across budgets. Third, some models neither improved nor declined monotonically: DeepSeek R1 stayed in the upper-40s to low-50s, and GLM-5 fluctuated between 37.7\% and 48.6\% without a clear trend in iteration count.

This suggests that additional iterations are not always necessarily beneficial. For stronger models that already answer correctly in a few steps, extra iterations introduce opportunities to second guess, experiment an unnecessary number of times, or accumulate context that leads to worse reasoning. For weaker models, additional iterations provide more chances to explore and converge. The implication is that $I_{\max}$ should be chosen per model rather than fixed across the benchmark.

\subsection{Qualitative vs. Quantitative Reasoning}
Consistent with our hypothesis in Section 3.6, the models performed significantly better on comparison and observation tasks compared to calculation and computation tasks.

Comparison problems, which ask which object satisfies a physical property, often demand only a single well-chosen experiment followed by qualitative interpretation. The models are able to do this even when the maximum iteration $I_{\max}$ is low.

Calculation and computation tasks, by contrast, require precise numerical reasoning and multiple tool invocations. Some common failure modes included applying the wrong physics formula, and using incorrect values extracted from the simulator. Sometimes, the LLMs even submitted intermediate steps as their answer instead of the actual final answers.

\begin{table*}[h]
\centering
\caption{Per-model metrics averaged across available iteration budgets.
         Accuracy is the fraction of scenes answered correctly.
         Ground. = Grounded-ness Score, Act. Val. = Action Validity Score.
         All models cover $I_{\max} \in \{5,10,15,20\}$.
         Full per-dimension breakdown in Appendix~\ref{app:dimensions}.}
\label{tab:main_results}
\renewcommand{\arraystretch}{0.90}
\begin{tabular}{lcccc}
\toprule
\textbf{Model} & \textbf{Acc.\ (\%)} & \textbf{Avg.\ FS} & \textbf{Ground.} & \textbf{Act.\ Val.} \\
\midrule
Claude Opus 4.7  & 40.7 & 60.0 & 0.544 & 0.999 \\
Gemini 3.1 Pro   & 49.5 & 58.4 & 0.598 & 1.000 \\
GPT-5.5          & 50.0 & 64.1 & 0.729 & 1.000 \\
DeepSeek R1      & 48.5 & 55.4 & 0.514 & 1.000 \\
GLM-5            & 45.2 & 50.0 & 0.744 & 1.000 \\
Kimi K-2.5       & 41.9 & 46.7 & 0.702 & 0.999 \\
Qwen 3.5         & 28.6 & 44.9 & 0.712 & 1.000 \\
\bottomrule
\end{tabular}
\end{table*}

\subsection{Common Failure Modes}

We identified six recurring failure modes: premature answer submission, insufficient simulation time, answer format mismatch, object ID confusion, wrong parameter substitution, and simulation trust vs. theoretical reasoning. Figure~\ref{fig:agent_loop} in Appendix~\ref{app:traces} illustrates the first of these on Scene 48, where the agent terminates after 2 of 5 available iterations by reading the incline angle off static geometry instead of experimentally locating the sliding threshold. Full interaction traces for each failure mode are provided in Appendix~\ref{app:traces}.

\subsection{Answer Normalization in Scoring}
\label{sec:normalization}

A recurring challenge in free-form benchmark evaluation is that an automatic matcher can mark a physically correct answer as wrong purely because of surface form. Similar free-form agentic benchmarks that suffer from this issue \citep{hendrycks2021measuring,mialon2023gaia,chandak2025answer} address it with equivalence-aware classification. PhysMent's default matcher is intentionally \emph{strict}: it applies the numerical tolerance of Eq.~\eqref{eq:correctness} and otherwise requires normalized substring containment. To separate genuine reasoning failures from matcher false negatives, we additionally define an \textbf{answer-normalization protocol} consisting of four documented, rule-based equivalence classes. A model response is regraded from Fail to Pass if and only if it falls into one of the following pre-registered classes verified against the scene's ground truth:

\begin{enumerate}[nosep,leftmargin=1.4em,label=\textbf{(\arabic*)}]
  \item \textbf{Equivalent object notation:} correct object under an alias the matcher missed (e.g.\ \texttt{"2"} vs \texttt{"object\_2"}). \emph{(15 cases, 5 scenes.)}
  \item \textbf{Trivial rounding:} correct formula, off by rounding beyond the tolerance of Eq.~\eqref{eq:correctness}. \emph{(35 cases, 10 scenes.)}
  \item \textbf{Conceptual description:} correct physical description equivalent to the keyword ground truth (e.g.\ ``slows, stops, and reverses'' for elastic collision). \emph{(63 cases, 11 scenes.)}
  \item \textbf{Partial answer format:} correct quantity embedded in extra or non-canonical formatting. \emph{(37 cases, 7 scenes.)}
\end{enumerate}
These rules account for $150$ re-grades over $33$ unique scenes. Two scenes dominate the conceptual class (Scenes 17 and 45 contribute 36 of 63), and Scene 52 contributes 17 of 37 partial-format cases.

\subsection{Limitations}

PhysMent has several limitations. The benchmark covers only classical mechanics and does not cover other physics topics. It is purely language-based and does not test visual perception of physical scenes. The iteration budget $I_{\max}$ imposes an artificial ceiling on experimental complexity. Our strict default matcher could be made more flexible through fuzzy matching or semantic-equivalence checking. Finally, our answer normalization protocol can be improved with additional equivalence classes.

\section{Future Work}
\label{sec:future}

\paragraph{Vision and embodiment.}
Because models frequently abandoned simulation after minimal tool calls (Section~\ref{sec:discussion}), a key open question is whether rendered RGB(D) observations alongside state readouts would improve temporal reasoning. Extending PhysMent with visual observations and vision language model interfaces would test perceptual grounding alongside tool-based reasoning.

\paragraph{Physics coverage and fidelity.}
Since all failure modes occurred within classical mechanics, it remains unknown whether the same patterns hold for fluids, deformation, or thermodynamics. Future versions could incorporate additional simulation substrates and calibration tasks closer to laboratory practice.

\paragraph{Models, budgets, and cost.}
Given that models such as GPT-5.5 and Claude Opus 4.7 showed marginal decline with more iterations while weaker models improved, adaptive stopping policies and think-mode ablations remain important open directions, as does standardizing the relationship between cost, latency, and accuracy reports across providers.

\paragraph{Dataset hygiene and community extensions.}
The 33 normalized scenes highlight the need for stricter answer format validation at construction time. Curating contributor guidelines and supporting community submitted scenes with automated validity checks would improve the long term benchmark stability.

\section{Conclusion}
\label{sec:conclusion}

We introduce PhysMent, a benchmark that tests how well language models reason about physics by interacting with a MuJoCo simulator. Unlike static benchmarks, PhysMent requires models to actively discover information through experimentation. Across 105 scenes and seven frontier models, we found that no model exceeded 67\% overall accuracy, despite category-level peaks above 80\%, and that increasing the iteration budget did not reliably improve performance, as some models plateaued, and others degraded with more iterations. These findings point toward the need for reasoning architectures that can plan experiments adaptively and maintain grounding across long multi-turn interactions, rather than simply extended context windows or additional tool calls.

\bibliography{physment}
\bibliographystyle{icml2026}

\appendix
\section{Structured Comparison of PhysMent to Key Benchmarks}
\label{app:comparison}
Table~\ref{tab:benchmark_comparison} compares PhysMent to current representative benchmarks on physics-reasoning. 

\begin{table*}[t]
\centering
\caption{Structural comparison of PhysMent with representative physical-reasoning benchmarks. \emph{Inter.}\ = interactive; \emph{Sim.}\ = grounded in a physics engine; \emph{Multi-turn} = sustained adaptive interaction (vs.\ single invocation); \emph{Proc.\ metrics} = reports process/trajectory metrics beyond correctness; \emph{Budget} = studies an iteration/interaction budget. ``--'' denotes not applicable.}
\label{tab:benchmark_comparison}
\small
\renewcommand{\arraystretch}{1.05}
\begin{tabular}{lccccccc}
\toprule
\textbf{Benchmark} & \textbf{LLM} & \textbf{Inter.} & \textbf{Sim.} & \textbf{Multi-turn} & \textbf{Proc.\ metrics} & \textbf{Budget} & \textbf{\# Tasks} \\
\midrule
PIQA \citep{bisk2020piqa}            & \checkmark & $\times$ & $\times$ & $\times$ & $\times$ & -- & $\sim$20k \\
GPQA \citep{rein2024gpqa}            & \checkmark & $\times$ & $\times$ & $\times$ & $\times$ & -- & 448 \\
PhysReason \citep{zhang2025physreason} & \checkmark & $\times$ & $\times$ & $\times$ & \checkmark & -- & 1200 \\
PHYBench \citep{qiu2025phybench}     & \checkmark & $\times$ & $\times$ & $\times$ & \checkmark & -- & 500 \\
PHYRE \citep{bakhtin2019phyre}       & $\times$   & \checkmark & \checkmark & $\times$ & $\times$ & -- & 50 templates \\
ScienceWorld \citep{wang2022scienceworld} & \checkmark & \checkmark & rule-based & \checkmark & $\times$ & -- & 30 tasks \\
Mind's Eye / UTOPIA \citep{liu2022minds} & \checkmark & \checkmark & \checkmark & $\times$ & $\times$ & -- & 39 subtasks \\
\midrule
\textbf{PhysMent (ours)}             & \checkmark & \checkmark & \checkmark & \checkmark & \checkmark & \checkmark & 105 scenes \\
\bottomrule
\end{tabular}
\end{table*}
\section{Per-Dimension Scores}
\label{app:dimensions}

\noindent Table~\ref{tab:per_dimension} reports per-model scores across all six scoring dimensions.

\paragraph{Why the $C$ column differs from Accuracy.}
The \textbf{Acc.} column is the strict pass/fail accuracy of Eq.~\eqref{eq:correctness} (a scene is either correct or not). The \textbf{Correctness dimension $C$} used inside the Final Score is a \emph{graded} $[0,1]$ quantity that awards partial credit on multi-part computation scenes (Section~\ref{sec:method}): a scene whose ground truth is a tuple (e.g., ``angular velocity \emph{and} rotational energy after 2\,s, then again after a second torque'') contributes the fraction of sub-answers within tolerance, rather than collapsing to $0$ on any single miss. This is why, for example, GPT-5.5 has $\text{Acc.}=50.0\%$ but $C=0.583$ (partial credit raises $C$ above the all-or-nothing pass rate), whereas GLM-5 has $\text{Acc.}=45.2\%$ but $C=0.351$ (it more often produces a correct \emph{final} value while missing intermediate sub-answers, so its graded score falls below its strict pass rate). The two columns are therefore expected to differ; they are not an inconsistency.
\begin{table*}[h]
\centering
\caption{Per-model scores across all six scoring dimensions, averaged over available 
         iteration budgets. $C$ = Correctness, $E$ = Efficiency, $G$ = Groundedness, 
         $V$ = Action Validity, $R$ = Reasoning Diversity, $G'$ = Generalization, 
         FS = Final Score (0--100). Final Score recomputed uniformly using fixed weights 
         $(w_C, w_E, w_G, w_V, w_R, w_{G'}) = (0.40, 0.15, 0.15, 0.10, 0.10, 0.10)$ 
         for comparability across runs.}
\label{tab:per_dimension}
\renewcommand{\arraystretch}{1.25}
\begin{tabular}{lcccccccc}
\toprule
\textbf{Model} & \textbf{Acc.\ (\%)} & $C$ & $E$ & $G$ & $V$ & $R$ & $G'$ & \textbf{FS} \\
\midrule
GPT-5.5                    & 50.0 & 0.583 & 0.537 & 0.729 & 1.000 & 0.701 & 0.476 & 64.1 \\
Gemini 3.1 Pro   & 49.5 & 0.555 & 0.420 & 0.598 & 1.000 & 0.657 & 0.437 & 58.4 \\
Claude Opus 4.7            & 40.7 & 0.503 & 0.655 & 0.544 & 0.999 & 0.613 & 0.580 & 60.0 \\
DeepSeek R1                & 48.5 & 0.480 & 0.418 & 0.514 & 1.000 & 0.633 & 0.588 & 55.4 \\
GLM-5                      & 45.2 & 0.351 & 0.238 & 0.744 & 1.000 & 0.700 & 0.426 & 50.0 \\
Kimi K-2.5                 & 41.9 & 0.308 & 0.201 & 0.702 & 0.999 & 0.708 & 0.381 & 46.7 \\
Qwen 3.5                   & 28.6 & 0.224 & 0.280 & 0.712 & 1.000 & 0.678 & 0.432 & 44.9 \\
\bottomrule
\end{tabular}
\end{table*}

\section{Scene Construction Guidelines}
\label{app:scenes}

Each scene is defined by two files under \texttt{Scenes/Scene\{N\}/} with matching filenames \texttt{scene\{N\}.json} and \texttt{scene\{N\}.xml} for $N\!\in\![7,111]$.

\paragraph{Operational category definitions.}
Table~\ref{tab:category_defs} gives the design-time rule used to assign each scene to a difficulty/modality cell, together with a representative scene. These are the rules referenced in Section~\ref{sec:method}; they are applied at construction time so that category membership is reproducible rather than subjective. 

\begin{figure*}[t]
\centering
\small
\begin{tikzpicture}[
  iter/.style={
    draw,
    rounded corners,
    minimum width=2.25cm,
    minimum height=1.05cm,
    align=center,
    font=\bfseries\small,
    fill=gray!8
  },
  trace/.style={
    draw,
    rounded corners,
    text width=10.8cm,
    minimum height=1.25cm,
    align=left,
    font=\small,
    inner sep=6pt
  },
  verdict/.style={
    draw,
    rounded corners,
    text width=13.6cm,
    align=left,
    font=\small,
    inner sep=7pt,
    fill=red!8
  },
  arr/.style={-{Stealth[length=2mm]}, thick}
]

\node[iter] (i1) {Iteration 1};
\node[trace, right=0.75cm of i1] (t1) {
\textbf{Agent reasoning:} ``I'll measure the incline angle directly.''\\
\textbf{Tool calls and returns:}
\texttt{get\_position(block)} $\rightarrow [-2.0,\,-0.6,\,0.5]$;\quad
\texttt{step(0.5)} $\rightarrow$ \texttt{simulation\_time = 0.5s}
};

\node[iter, below=0.8cm of i1] (i2) {Iteration 2};
\node[trace, right=0.75cm of i2] (t2) {
\textbf{Agent reasoning:} ``From geometry: $\mu = \tan(\theta) \approx 0.24$.''\\
\textbf{Terminal call:}
\texttt{answer(0.24)} $\rightarrow$ episode terminates after 2 of 5 available iterations.
};

\node[verdict, below=0.9cm of i2, xshift=5.85cm] (v) {
\textbf{$\times$ Premature answer submission.}
The agent derived $\mu$ from static geometry and submitted an answer without varying the incline angle or detecting the sliding threshold.
};

\draw[arr] (i1) -- (i2);
\draw[arr] (t1) -- (t2);
\draw[arr] (i2) -- (v);

\end{tikzpicture}
\caption{One PhysMent episode from Scene 48. The agent submits its answer after only 2 of 5 available iterations, illustrating premature answer submission: it derives $\mu$ from static geometry rather than experimentally varying the incline angle to detect sliding onset. See Appendix~\ref{app:traces} for the full trace.}
\label{fig:agent_loop}
\end{figure*}
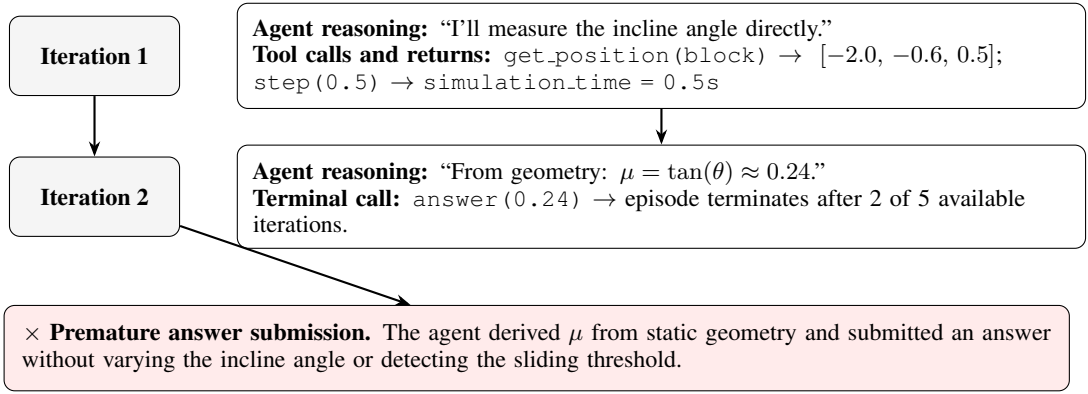

\begin{table*}[!t]
\centering
\caption{Operational definitions of the PhysMent difficulty/modality axes. \emph{Ref.\ steps} = number of distinct simulator-intervention steps in the reference solution used to assign Easy ($\leq 3$) vs.\ Hard ($>3$). \emph{Concepts} = number of distinct physical concepts required (Single $=1$, Multi $\geq 2$).}
\label{tab:category_defs}
\small
\renewcommand{\arraystretch}{1.15}
\begin{tabular}{p{3.1cm}ccp{4.3cm}p{3.2cm}}
\toprule
\textbf{Category} & \textbf{Concepts} & \textbf{Ref.\ steps} & \textbf{Modality rule} & \textbf{Representative scene} \\
\midrule
Easy Single Phys.     & 1 & $\leq 3$ & standard & free-fall velocity readout (Scene 7) \\
Easy Single + Obj.C.  & 1 & $\leq 3$ & must \texttt{create\_objects} for a reference body & add reference mass (Scene 16) \\
Easy Single + Hid.    & 1 & $\leq 3$ & must \texttt{find\_objects} to reveal a masked body & hidden ball discovery (Scene 21) \\
Hard Single Phys.     & 1 & $>3$ & standard, adaptive loop required & friction threshold via angle search (Scene 26) \\
Hard Single + Obj.C.  & 1 & $>3$ & object creation + adaptive loop & Scene 41 \\
Hard Single + Hid.    & 1 & $>3$ & hidden object + adaptive loop & Scene 46 \\
Easy Multi            & $\geq 2$ & $\leq 3$ & standard & combined kinematics+energy (Scene 51) \\
Easy Multi + Obj.C.   & $\geq 2$ & $\leq 3$ & object creation & Scene 66 \\
Easy Multi + Hid.     & $\geq 2$ & $\leq 3$ & hidden object & Scene 71 \\
Hard Multi            & $\geq 2$ & $>3$ & standard, adaptive loop & Scene 76 \\
Hard Multi + Obj.C.   & $\geq 2$ & $>3$ & object creation + adaptive loop & Scene 91 \\
Hard Multi + Hid.     & $\geq 2$ & $>3$ & hidden object + adaptive loop & Scene 96 \\
Create / Del.\ / Att. & -- & varies & scene-graph manipulation is the task & attach two bodies (Scene 101) \\
\bottomrule
\end{tabular}
\end{table*}

\paragraph{Metadata file (\texttt{scene\{N\}.json}).}
Specifies: scene name, task, and \texttt{problem\_type}; ground-truth
\texttt{answer}; \texttt{expected\_behavior}; \texttt{reasoning};
object list with IDs; and \texttt{object\_permissions} per attribute.

\noindent\textbf{MJCF file (\texttt{scene\{N\}.xml}).}
A valid MuJoCo model. Each dynamic object requires a
\texttt{<body name="object\_\{id\}">} wrapper, a \texttt{<geom>}
defining shape, size, density, and friction, and
\texttt{<joint name="object\_\{id\}\_joint" type="free"/>} for
unconstrained rigid-body motion.

\section{Complete Tool API}
\label{app:tools}

\begin{table*}[h]
\centering
\caption{All 26 simulation tools exposed to agents in PhysMent.}
\label{tab:tools}
\small
\begin{tabular}{lp{5.5cm}}
\toprule
\textbf{Tool} & \textbf{Description} \\
\midrule
\texttt{step(dur)}                        & Advance simulation by \texttt{dur} seconds \\
\texttt{get\_position(id)}                & Return $[x,y,z]$ and current simulation time \\
\texttt{move\_object(id, x, y, z)}        & Teleport object to absolute position \\
\texttt{change\_position(id, d, fr)}      & Translate by $[dx,dy,dz]$ in world or local frame \\
\texttt{get\_velocity(id)}                & Return $[v_x,v_y,v_z]$ linear velocity \\
\texttt{set\_velocity(id, v)}             & Set linear velocity vector \\
\texttt{get\_acceleration(id)}            & Estimate acceleration from velocity difference \\
\texttt{apply\_force(id, f)}              & Apply force $[f_x,f_y,f_z]$ \\
\texttt{apply\_torque(id, $\tau$)}        & Apply torque $[\tau_x,\tau_y,\tau_z]$ \\
\texttt{get\_torque(id)}                  & Return currently applied torque \\
\texttt{get\_parameters(id)}              & Return mass, bounding box, geometry type \\
\texttt{get\_displacement(id)}            & Euclidean distance from initial position \\
\texttt{compute\_force(id, m)}            & Compute $F = ma$ \\
\texttt{get\_kinetic\_energy(id, m)}      & Compute $\tfrac{1}{2}mv^2$ \\
\texttt{get\_potential\_energy(id, m, g)} & Compute $mgh$ \\
\texttt{get\_rotational\_energy(id, m)}   & Compute $\tfrac{1}{2}I\omega^2$ \\
\texttt{get\_momentum(id, m)}             & Compute $p = mv$ \\
\texttt{get\_angular\_momentum(id, m)}    & Compute $L = I\omega$ \\
\texttt{get\_center\_of\_mass()}          & Scene-wide center of mass \\
\texttt{detect\_collision(id1, id2)}      & Check contact; apply elastic response \\
\texttt{quat\_to\_rot\_matrix(q)}         & Quaternion to $3{\times}3$ rotation matrix \\
\texttt{create\_objects(n, p, $\rho$, c)} & Add sphere (modifies XML, reloads) \\
\texttt{delete\_objects(id)}              & Remove object (modifies XML, reloads) \\
\texttt{find\_objects()}                  & Reveal hidden bodies by resetting colors \\
\texttt{attach\_objects(id1, id2)}        & Create joint between two bodies \\
\texttt{answer(ans)}                      & Submit final answer; terminate episode \\
\bottomrule
\end{tabular}
\end{table*}

Table~\ref{tab:tools} lists all 26 simulation tools exposed to agents, organized by function category as described in Section~\ref{sec:method}.

\section{Prompting Strategies}
\label{app:prompting}

\textbf{Zero-shot.} The prompt contains the task description, visible
object attributes (filtered by $\bP_i$), full tool API documentation,
coordinate-frame conventions, and answer-format instructions. No
worked examples are provided.

\textbf{One-shot / few-shot CoT.} One or three worked examples are attached to the prompt: a dropped ball, a pushed box, and a sphere on a ramp, each with a reasoning trace and tool calls showing how to interact with the simulator.

\section{Failure Mode Interaction Traces}
\label{app:traces}

This appendix presents the raw prompts and model responses for the six failure modes discussed in Section~\ref{sec:discussion}. Each example shows the initial task prompt, key model responses, submitted answer, and correct answer.

\subsection*{Scene 48: Premature Answer Submission}
\textbf{Model:} DeepSeekAgent $|$ \textbf{Budget:} $I_{\max}=5$

\textbf{Prompt:} \textit{Determine the coefficient of static friction using the critical angle at which the block begins to slide.} Objects: block at position $[-2.0, -0.6, 0.5]$, ramp at origin.

\textbf{Iteration 1 response:} The model computed the incline angle from static block coordinates: rise $= 0.5$ m, run $= 2.0$ m, $\mu_s = \tan(\theta) = 0.25$. It then stepped the simulation once for $0.5$ s to ``verify''.

\textbf{Iteration 2 (final):} After observing displacement, the model refined the geometry calculation: run $= \sqrt{(-2.0)^2 + (-0.6)^2} \approx 2.088$ m, submitted $\mu_s = 0.240$.

\textbf{Submitted:} $0.24$ \quad \textbf{Correct:} $\mu_s = \tan(\theta_c)$ via experiment.

\textit{The model never varied the incline angle or detected the onset of sliding. It derived $\mu$ purely from static geometry in 2 of 5 available iterations.}

\subsection*{Scene 89: Insufficient Simulation Time}
\textbf{Model:} QwenAgent $|$ \textbf{Budget:} $I_{\max}=15$

\textbf{Prompt:} \textit{Determine the period of oscillation for a simple pendulum with a length of 2 meters.}

\textbf{Iteration 1--6:} The model attempted to displace the pendulum via \texttt{move\_object} (shape error), \texttt{apply\_force}, and \texttt{step(0.1)}. After $0.1$ s of simulation, position and velocity remained $[0,0,0]$ — the pendulum did not move.

\textbf{Iteration 7 (final):} Seeing no oscillation, the model abandoned simulation and applied the formula: $T = 2\pi\sqrt{L/g} = 2\pi\sqrt{2/9.81} \approx 2.843$ s.

\textbf{Submitted:} $2.843$ s \quad \textbf{Correct:} $2.837$ s.

\textit{Only $0.1$ s of simulation was run, which is $3.5\%$ of one full period, and the pendulum never oscillated due to a failed displacement step. The model abandoned the simulator and fell back to the textbook closed-form formula rather than measuring the period experimentally. Although its hand-computed value happens to fall within the $5\%$ tolerance of Eq.~\eqref{eq:correctness}, the episode exemplifies the failure mode this category targets—abandoning the experimental loop—rather than a numerical scoring error.}

\subsection*{Scene 63: Answer Format Mismatch}
\textbf{Model:} GLMAgent $|$ \textbf{Budget:} $I_{\max}=20$

\textbf{Prompt:} \textit{Observe how angular velocity changes when masses are moved closer to the center.} Objects: rotating platform, two masses at $r=0.4$ m.

\textbf{Iterations 1--18:} The model made 45 tool calls over 18 iterations, attempting to spin the platform via \texttt{apply\_torque}, \texttt{apply\_force}, and \texttt{set\_velocity}. All attempts produced near-zero angular momentum due to scene configuration issues.

\textbf{Final response:} The model fell back to theory, computing $I_{\text{initial}} = 0.00271$ kg$\cdot$m$^2$, $I_{\text{final}} = 0.00258$ kg$\cdot$m$^2$, and $\omega_{\text{final}}/\omega_{\text{initial}} = 1.05$. Submitted the ratio $1.05$.

\textbf{Submitted:} $1.05$ \quad \textbf{Correct:} $2$ (object ID of the faster-spinning mass).

\textit{The scene expected a comparison object ID, not a numerical ratio. The model's physics was correct but it misread the answer format entirely.}

\subsection*{Scene 68: Object ID Confusion}
\textbf{Model:} GLMAgent $|$ \textbf{Budget:} $I_{\max}=20$

\textbf{Prompt:} \textit{Identify which rope has higher tension based on the mass it supports.} Objects: \texttt{object\_1} (mass\_1, density 1.0), \texttt{object\_2} (mass\_2, density 2.0), \texttt{object\_3} (rope\_1), \texttt{object\_4} (rope\_2).

\textbf{Iterations 1--7:} The model correctly measured masses ($0.00419$ kg vs $0.00838$ kg), identified rope\_2 above mass\_2, and concluded rope\_2 has twice the tension.

\textbf{Final response:} ``rope\_2 (\texttt{object\_4}) has higher tension.'' Submitted ID \texttt{4}.

\textbf{Submitted:} $4$ (rope\_2) \quad \textbf{Correct:} $2$ (mass\_2).

\textit{The answer key expected the ID of the heavier mass, not the rope. The model's physics reasoning was entirely correct but it submitted the rope's ID instead of the mass's ID.}

\subsection*{Scene 91: Wrong Parameter Substitution}
\textbf{Model:} GPT-5.5 $|$ \textbf{Budget:} $I_{\max}=15$

\textbf{Prompt:} \textit{Determine the rotational kinetic energy of a disk with radius $0.5$ m, mass $2$ kg, and angular velocity $10$ rad/s.}

\textbf{Iteration 1:} The model called \texttt{get\_parameters} and received simulator mass $= \pi \approx 3.14159$ kg (computed internally from geometry and density).

\textbf{Iteration 2 (final):} ``The simulator shows the actual disk mass is $3.14159$ kg, so I should use that verified value.'' Applied $KE = \frac{1}{4}mr^2\omega^2$ with $m = \pi$: result $= 19.635$ J.

\textbf{Submitted:} $19.635$ J \quad \textbf{Correct:} $12.5$ J ($= \frac{1}{4} \times 2 \times 0.25 \times 100$).

\textit{The formula structure was correct but the model explicitly discarded the problem's stated mass in favour of the simulator's geometry-derived mass, treating the simulator as more authoritative than the problem parameters.}

\subsection*{Scene 14: Simulation Trust vs.\ Theoretical Reasoning}
\textbf{Model:} KimiAgent $|$ \textbf{Budget:} $I_{\max}=10$

\textbf{Prompt:} \textit{Determine which block will travel further when pushed with the same force.} Objects: \texttt{object\_1} (low friction), \texttt{object\_2} (high friction), equal mass.

\textbf{Iterations 1--9:} The model applied equal forces to both blocks and stepped for $2$ s. Measured displacements: \texttt{object\_1} $= 0.100$ m, \texttt{object\_2} $= 4.001$ m — a $40\times$ difference inconsistent with friction physics.

\textbf{Final response:} ``Physically, the low friction block should travel further \ldots However, my simulation results indicate that \texttt{object\_2} (high friction) traveled significantly further.'' Submitted \texttt{object\_2}.

\textbf{Submitted:} $2$ (high friction) \quad \textbf{Correct:} $1$ (low friction).

\textit{The model explicitly identified the physical contradiction in its own reasoning but deferred to the anomalous simulator output. The displacement anomaly was caused by incorrectly initialized friction coefficients in the scene XML.}

\section{Examples of JSON/XML Files, Initial Scene Prompt for Experiment Logs}
\paragraph{Example Scene JSON and XML Specifications:}
We present the corresponding JSON metadata file and XML scene specification for the example rotation and kinetic-energy task. The JSON file defines the task-level structure exposed to the evaluation pipeline, including the scene name, task description, problem type, ground-truth answer, expected behavior, object list, and object-level permissions. The XML file encodes the underlying MuJoCo scene, specifying the physical object geometry, pose, density, and joint configuration used by the simulator during tool-mediated experimentation.

\begin{promptblock}[JSON Metadata: Scene-Level Task Definition]{yellow}
{
  "metadata": {
    "scene_name": "Rotation and Kinetic Energy",
    "task": "A disk is set in motion by applying a torque of 20 N $\cdot$ m. First, calculate its angular velocity and rotational energy after 2 seconds of simulation. Then, apply a torque of 10 N*m and calculate the new angular velocity and rotational energy after 2 seconds.",
    "problem_type": "computation"
  },
  "answer": "4.0, 79.6, 6.0, 179.0",
  "expected_behavior": "The LLM should apply a torque to rotate the object and calculate the angular velocity and rotational energy after 2 seconds of simulation.",
  "reasoning": "By applying torque, the object will rotate and accumulate angular momentum. The rotational energy can be calculated using the formula for kinetic energy of rotation.",
  "number_of_objects": "1"
}
\end{promptblock}

\begin{promptblock}[JSON Object Registry]{cyan}
{
  "objects": {
    "object_1": {
      "name": "disk",
      "object_id": "1"
    }
  }
}
\end{promptblock}

\begin{promptblock}[JSON Object-Permission Mask]{green}
{
  "object_permissions": {
    "object_1_permissions": {
      "geom_type": true,
      "geom_density": true,
      "body_mass": true,
      "geom_size": true,
      "geom_radius": true,
      "body_pos": true,
      "body_quat": true
    }
  }
}
\end{promptblock}

\begin{promptblock}[XML Scene Header and Global Physics Options]{blue}
<mujoco model="rotation_kinetic_energy_test">
    <option gravity="0 0 -9.81"/>
    <worldbody>
\end{promptblock}

\begin{promptblock}[XML Dynamic Object Definition]{orange}
        <!-- Disk -->
        <body name="object_1" pos="0 0 1">
            <geom type="cylinder" size="0.2 0.05" density="40000" rgba="0 1 0 1"/>
            <joint name="object_1_joint" type="free"/>
        </body>
\end{promptblock}

\begin{promptblock}[XML Scene Closure]{purple}
    </worldbody>
</mujoco>
\end{promptblock}

\paragraph{Specification structure:}
The JSON file separates semantic task information from simulator state. The \texttt{metadata} field specifies the natural-language task, scene name, and problem type; the \texttt{answer} field stores the target output used for evaluation; and the \texttt{expected\_behavior} and \texttt{reasoning} fields document the intended experimental strategy. The object registry maps symbolic object identifiers, such as \texttt{object\_1}, to named physical entities, while the permission mask determines which attributes are exposed to the LLM through the prompt and tool interface.

\paragraph{Simulator encoding:}
The XML file provides the executable MuJoCo scene loaded by the evaluation runtime. In this example, the scene contains a single disk represented as a cylindrical geometry with radius and half-height specified by \texttt{size="0.2 0.05"}. The body is initialized at position \texttt{pos="0 0 1"} and assigned a free joint, allowing the simulator to update its translational and rotational state under applied forces and torques. This separation between JSON metadata and XML physics specification allows the benchmark to expose controlled task information to the LLM while preserving a grounded simulation backend for tool execution.
\paragraph{Initial Scene Prompt for Experiment Logs:} We present an example initial scene prompt given to the LLM at the beginning of an experimental episode. The prompt provides task-level metadata extracted from the scene JSON file, while the associated XML scene specification is loaded by the evaluation pipeline to enable valid simulator interactions through tool calls. To improve readability, the prompt is divided into the following components: Scene Description and Task, Available Objects and Parameters, Tool API Listing, Tool Return Format and Answer Submission, Example Assistant--Environment Interaction (for one-shot or few-shot chain-of-thought prompting), Answer Format Requirements, Final Submission Guidelines, and JSON Tool-Call Formatting Rules.

\begin{promptblock}[Scene Description and Task]{yellow}
Scene Description: Rotation and Kinetic Energy.

Task: A disk is set in motion by applying a torque of 20 N*m. First, calculate its angular velocity and rotational energy after 2 seconds of simulation. Then, apply a torque of 10 N*m and calculate the new angular velocity and rotational energy after 2 seconds.
\end{promptblock}

\begin{promptblock}[Available Objects and Parameters]{cyan}
Available Objects and Parameters:
Object id: object_1, Object name: disk

Quaternion: A quaternion is used to represent rotation in 3D space. The four numbers represent rotations along the X, Y, Z axes and a scalar component.

If an attribute has 'n/a' right beside it, that means you CANNOT access that attribute's value, so keep that in mind when running through the experiment.
\end{promptblock}

\begin{promptblock}[Tool API Listing]{green}
You may use the following tools along with their description to interact with the scene. These functions accept parameters given below, and return data or perform simulation updates:

[
  {
    "name": "step",
    "description": "keeps on moving the simulator forward in time",
    "arguments": {"duration": "float"},
    "return type": {"results": null}
  },
  {
    "name": "apply_force",
    "description": "applies a force vector to an object",
    "arguments": {"object_id": "str", "force_vector": "list[float]"},
    "return type": {"status": "str", "object_id": "int", "force": "list[float]"}
  },
  {
    "name": "get_velocity",
    "description": "retrieves the velocity vector of an object",
    "arguments": {"object_id": "str"},
    "return type": {"velocity": "array"}
  },
  ...
  {
    "name": "answer",
    "description": "submits an answer back to the system for checking or logging",
    "arguments": {"answer": "str or float"},
    "return type": {"acknowledged": "bool"}
  }
]
\end{promptblock}

\begin{promptblock}[Tool Return Format and Answer Submission]{blue}
Every time you call a tool, you will receive a dictionary containing the outputs. For example, if you call `get_velocity` on `object_1`, the return might be:
{"vx": 0.0, "vy": -3.2, "vz": 0.0}

You only have **one chance** to answer the question. When you're confident, submit your final answer using:
{"tool": "answer", "parameters": {"answer": "<your_answer>"}}
\end{promptblock}

\begin{promptblock}[Example Assistant--Environment Interaction]{orange}
<THIS IS AN EXAMPLE PROBLEM OF THE INPUTS(ASSISTANT) AND OUTPUTS(ENVIRONMENT) THAT SHOULD TAKE PLACE>

Problem: You are given a ball and a ground surface for reference. Drop the ball from a height of 10 units and figure out the velocity of the object after 0.5 seconds.

<assistant>
I see that I have to move the ball up 10 units so I will do that.
json
[{"tool": "move_object", "parameters": {"object_id": "object_1", "x": 0, "y": 0, "z": 10}}]

<environment>
Results: [{"tool": "move_object", "parameters": {...}, "result": {"position": [0, 0, 10]}, "sim_time": 0}] What will you do next

<assistant>
Now I will simulate by using the step function to go 0.5 seconds forward.
json
[{"tool": "step", "parameters": {"duration": 0.5}}]

<environment>
Results: [{"tool": "step", "parameters": {...}, "result": null, "sim_time": 0.5}] What will you do next

<assistant>
Now I will use the get velocity function to figure out what I should output as my answer.
json
[{"tool": "get_velocity", "parameters": {"object_id": "object_1"}}]

<environment>
Results: [{"tool": "get_velocity", "parameters": {...}, "result": {"velocity": [0, 0, -4.9]}, "sim_time": 0.5}] What will you do next

<assistant>
Now I will call back the answer.
json
[{"tool": "answer", "parameters": {"answer": "-4.9"}}]

<END EXAMPLE>
\end{promptblock}

\begin{promptblock}[Answer Format Requirements]{purple}
Since the problem is a computation problem, your answer should be the calculated number that satisfies the task rounded to the nearest thousandths place if applicable.
\end{promptblock}

\begin{promptblock}[Final Simulation Guidelines]{red}
***FINAL GUIDELINES***

You must call `step` to simulate time progression.

Do not make any assumptions about the positions or states of objects; if you are unsure, you can use tools to get this information.

The z plane is the vertical plane, and the x and y planes are the horizontal planes.

Understand that you can use the tools to gather necessary information pertaining to all objects in the scene, not just the one you are trying to analyze.

If your json format is incorrect, the environment will tell you and the simulator will remain in the same state. If one of your tool calls has incorrect formatting, the previous tool calls will successfully execute but the incorrect tool and subsequent tools will not execute.
\end{promptblock}

\begin{promptblock}[JSON Tool-Call Formatting Rules]{magenta}
Remember that YOU MUST PROVIDE YOUR REASONING FOR EVERY ACTION you take, and then make sure to add a valid JSON format of an array of tool calls.

***When you are trying to call functions, use a string that goes as object_{object_id} for the object id, and use the name of the function as the tool name.***

For example, if you were to try and call the function get_velocity on object_1, you would write it as:
json
[{"tool": "get_velocity", "parameters": {"object_id": "object_1"}}]

RATHER THAN WRITING IT AS:
json
[{"tool": "get_velocity", "parameters": {"object_id": 1}}]

Remember that when giving multiple tool calls, you must give them in ONE SINGLE GO, as shown below.
json
[{"tool": "get_velocity", "parameters": {"object_id": "object_1"}},
 {"tool": "get_velocity", "parameters": {"object_id": "object_2"}}]

DO NOT GIVE MORE THAN ONE SET OF TOOL CALLS AT A TIME, OR THE SIMULATOR WILL NOT WORK, AND END UP GIVING YOU AN ERROR FOR JSON SYNTAX!!!
\end{promptblock}

\end{document}